\pdfoutput=1

\documentclass[11pt]{article}

\usepackage{acl}

\usepackage{times}
\usepackage{latexsym}
\usepackage{hyperref}
\usepackage{url}
\usepackage{enumitem}
\usepackage{array}
\usepackage{balance}
\usepackage{color}
\usepackage[toc,page]{appendix}
\usepackage{bbm}
\usepackage{algorithm}
\usepackage[noend]{algpseudocode} 
\usepackage{bbding}
\usepackage[noadjust,nosort]{cite}
\usepackage{booktabs}
\usepackage[normalem]{ulem}
\usepackage{setspace}
\usepackage{stfloats}
\usepackage{subcaption}
\usepackage{epsfig}
\usepackage{amsmath}
\usepackage{amssymb}
\usepackage{comment}
\usepackage{multirow}
\usepackage{wrapfig}
\usepackage{comment}

\usepackage[T1]{fontenc}

\usepackage[utf8]{inputenc}

\usepackage{microtype}

\title{E-LANG: Energy-Based Joint Inferencing\\ of Super and Swift Language Models}

 \author{{Mohammad Akbari, Amin Banitalebi-Dehkordi, Yong Zhang}
 \\ Huawei Technologies Canada Co., Ltd.
 \\ \texttt{\{mohammad.akbari, amin.banitalebi, yong.zhang3\}@huawei.com}
 }

\begin{document}
\maketitle


\begin{abstract}
Building huge and highly capable language models has been a trend in the past years. Despite their great performance, they incur high computational cost. A common solution is to apply model compression or choose light-weight architectures, which often need a separate fixed-size model for each desirable computational budget, and may lose performance in case of heavy compression. This paper proposes an effective dynamic inference approach, called E-LANG, which distributes the inference between large accurate Super-models and light-weight Swift models. To this end, a decision making module routes the inputs to Super or Swift models based on the energy characteristics of the representations in the latent space. This method is easily adoptable and architecture agnostic. As such, it can be applied to black-box pre-trained models without a need for architectural manipulations, reassembling of modules, or re-training. Unlike existing methods that are only applicable to encoder-only backbones and classification tasks, our method also works for encoder-decoder structures and sequence-to-sequence tasks such as translation. The E-LANG performance is verified through a set of experiments with T5 and BERT backbones on GLUE, SuperGLUE, and WMT. In particular, we outperform T5-11B with an average computations speed-up of 3.3$\times$ on GLUE and 2.9$\times$ on SuperGLUE. We also achieve BERT-based SOTA on GLUE with 3.2$\times$ less computations. Code and demo are available \href{https://developer.huaweicloud.com/develop/aigallery/notebook/detail?id=64199726-9aaf-4905-8f6f-4cae290df874}{here}.
\end{abstract}

\section{Introduction}
\label{sec:introduction}
With the introduction of influential language models such as BERT \citep{bert}, a trend in natural language processing (NLP) research has been to develop high capacity models and push their performance to new levels. Consequently, state-of-the-art (SOTA) results were achieved on various benchmarks using these models; GPT-3 \citep{gpt3}, XLNet \citep{xlnet}, RoBERTa \citep{roberta}, T5 \citep{t5}, ELECTRA \citep{electra}, and DeBERTa \citep{deberta} to name a few. 
A potential down-side, however, is that the number of parameters or floating point operations (FLOPs) for these models can get extremely large. For example, Gshard \citep{gshard} comes with 600B parameters with an enormous amount of computation. This in turn results in a higher inference latency, which is not desirable for latency-sensitive applications. 

A common solution to speed-up the large language models is to apply model compression \citep{gupta2020compression}. Although generally successful, compression does come with a trade-off on accuracy, and may lose performance if compression is heavy. In addition, these methods usually compress a model to a fixed smaller size, where a separate model is required for each possible computational budget. An alternative approach explored in the literature is to leverage dynamic inferencing in a way that examples may be routed to different (potentially lower cost) paths throughout the network. For example, a temporal early-exit model \citep{reasonet,yu2018fast} terminates the procedure of reading the input sequence when sufficient evidence has been found for accurate predictions. Instance-wise early-exiting \citep{deebert} is another technique, which allows a sample to adaptively choose from multiple available exit nodes if some conditions are met. Consequently, earlier exists require less computation and lead to a lower latency. Adjusting the size of the model at the inference time by choosing adaptive width and depth is also another approach employed for dynamic inference \citep{lengthadaptive,dynabert}. There is a variety of adaptive/dynamic inference approaches proposed, however, a general down-side for many of these methods is that often times they require a careful architecture design, manipulation of network modules, or even re-training.




In this paper, we propose a simple but rather effective approach of dynamically distributing the inference between the original large model (called the \textbf{Super} model) and a light-weight (e.g., compressed) model referred to as the \textbf{Swift} model. To this end, we design an energy-based decision making module that routes examples to the appropriate model based on the negative free energy of the latent space representations, such that the Swift model attains a high accuracy on the examples sent to it. The remaining samples are then forwarded to the Super model that is supposed to have a good performance on all examples. Since the Swift model can make highly accurate predictions over the majority of the samples, E-LANG significantly reduces the overall computational cost, while maintains the high accuracy of the Super model. Although simple, this strategy achieves SOTA results on multiple structures (e.g., T5 and BERT) and benchmarks (e.g., GLUE and SuperGLUE). Due to its desirable practical characteristics, this method is a strong candidate for the practical application of Super models.
The main contributions of the paper are as follows:
\begin{itemize}[leftmargin=*]
    \item Combining Super models with high accuracy and latency and Swift models with lower accuracy and latency, to achieve \textbf{high accuracy and low latency}. In other words, by employing our method, we can achieve the high levels of accuracy provided by Super models, but at a lower computational cost. Our method is easily adoptable, architecture agnostic, and orthogonal to many other existing methods. It can be applied to black-box pre-trained models without a need for architectural manipulations, careful reassembling of modules, or re-training. 
    \item An \textbf{energy-based routing mechanism} for directing examples to the Super or Swift. This provides a dynamic trade-off between the accuracy and computational cost that outperforms the previous works in both fixed-size and dynamic inference (with zero overhead for real-time adjustment of speed/accuracy). As such, E-LANG acts like a knob for adjusting the accuracy-latency trade-off in real-time during model serving.
    \item To the best of our knowledge, our method is the first generic approach to apply dynamic inference on both encoder-only and \textbf{encoder-decoder architectures} (e.g., T5) and also can extend the usage beyond classification tasks, to \textbf{sequence-to-sequence} tasks such as translation.
\end{itemize}


\section{Related Works}
\label{sec:related_works}
As mentioned, compression is a widely used strategy to speed-up the large language models \citep{gupta2020compression, gupta2020compression2}. This involves incorporating techniques such as quantization of weights and activations \citep{binarybert, qbert, kim2021bert, ternarybert, jin2021kdlsq}, knowledge distillation (KD) \citep{hinton2015distilling, tinybert, distilbert}, pruning/sharing \citep{gordon2020compressing, chen2020lottery}, multi-device distribution \citep{autosplit}, or a combination of these techniques \citep{cheng2017survey,polino2018model}.

Among all the compression techniques, creating a fixed-size small version of large models along with distillation has been popular in the recent years. \citet{distilbert} introduced DistillBERT, which was a smaller version of BERT trained with distillation for general purposes. Another compact variant of BERT was proposed by MobileBERT \citep{mobilebert} in which inverted bottleneck structures and progressive knowledge transfer were used. TinyBERT \citep{tinybert} also presented a novel two-stage transformer distillation for both pre-training and task-specific fine-tuning. In \citep{squeezebert}, the usage of grouped convolutions was studied to design SqueezeBERT. ELM \citep{elm}, a layer mapping search framework, was also proposed for improving downstream BERT distillation. A recent method, GhostBERT \citep{ghostbert}, employed softmax-normalized 1D convolutions as ghost modules to generate more features with cheap operations. 


Although compression techniques in general are effective, they come with a trade-off on accuracy, and may lose performance in case of high ratio compression. In addition, an individual fixed-size model is required for each possible computational budget. As stated in the introduction, the alternative solution is dynamic inference, which can be achieved with either early-exit or length/depth-adaptive models. One of the first temporal early-exit strategies was proposed by ReasoNet \citep{reasonet}, which stops its reading procedure when sufficient evidence has been found for answering a question. Similarly, in \citep{yu2018fast}, an early stopping method applicable to classification tasks was presented. DeeBERT \citep{deebert} also proposed an instance-wise multi-exit method via the entropy of the output probability distribution to speed-up BERT inference.

As a length-adaptive method, \citet{lengthadaptive} introduced a dynamic inference framework with one-shot training of transformers for both sequence- and token-level classification. Also, in \citep{dynabert}, an architecture named DynaBERT was proposed for adaptively adjusting the computations by choosing sub-networks of different widths and depths. Both Length-Adaptive and DynaBERT utilized knowledge distillation and data augmentation to improve their performance.

Although early-exit and adaptive methods have made significant progress and work well in practice, they often require architectural manipulation and re-training. In addition, they are only applicable to encoder-only backbones and classification tasks. In contrast, our method can work with out-of-the-box pre-trained models without a need for re-training and are also applicable for encoder-decoder structures and sequence-to-sequence tasks.

\begin{figure*}
    \centering
    \includegraphics[width=0.97\linewidth]{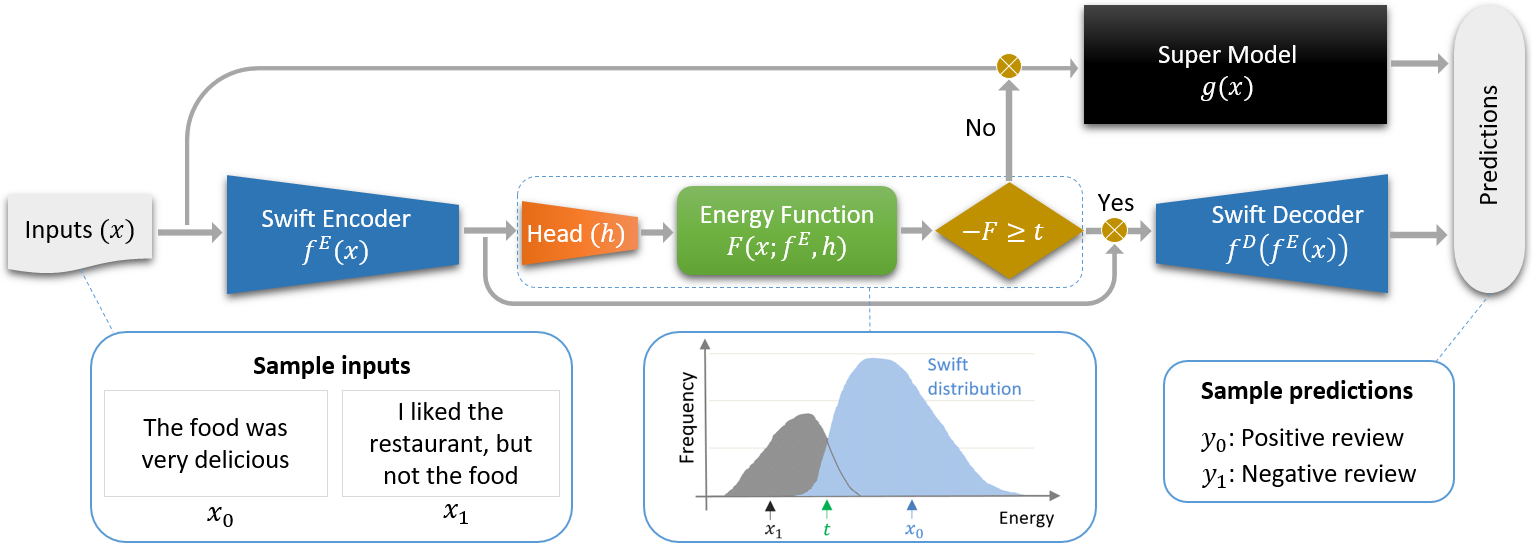}
    \vspace{-5pt}
    \caption{\label{fig:overall_framework} Overall framework of the proposed energy-based joint inference strategy (E-LANG).}
\end{figure*}

\section{Proposed Method}
\label{sec:proposed_method}
We propose a new energy-based joint inference method called E-LANG, where a large/accurate language model (\textbf{Super}) is jointly employed with a small/fast one (\textbf{Swift}) to achieve efficient inference without sacrificing the accuracy. To this end, inspired by the method in \citep{akbariebjr}, a routing mechanism empowered by energy-based models (EBM) is introduced to dynamically distribute the input samples between the Super and Swift models. Similar to the out-of-distribution (OOD) detection problem, our goal is to identify the OOD samples that are hard to handle for the Swift and forward them to the Super model. On the other hand, we have the in-distribution data for which the Swift can make highly reliable and accurate predictions. In other words, the routing mechanism needs to detect whether or not the input data fits in the Swift's distribution (i.e., the one the Swift has been trained with). Inspired by the success of EBMs in dealing with OOD detection problems \citep{lee2019energy}, the energy characteristics of data samples for an efficient and effective routing are investigated in our work. The overall framework of E-LANG is shown in Figure \ref{fig:overall_framework}.

\subsection{Energy-Based Models}
\label{ssec:energy_definition}
The goal of EBM is to build an energy function denoted by $E(\textbf{x}): \mathbb{R}^D \rightarrow \mathbb{R}$ that maps an input data $\textbf{x} \in \mathbb{R}^D$ to a non-probabilistic energy value $y \in \mathbb{R}$. To turn a collection of arbitrary energies for all possible outputs (denoted by $Y$) into a normalized probability distribution, Gibbs distribution can be used as follows \citep{lecun2006tutorial}:
\begin{equation}
    p(y|\textbf{x}) = \frac{e^{-E(\textbf{x},y)}}{\int_{y'\in Y} e^{-E(\textbf{x},y')}},
\label{eq:gibbs_dist}
\end{equation}
where the negative log of the denominator expresses the Helmholtz free energy \citep{lecun2006tutorial} defined as $F(\textbf{x}) = -log \big(\int_{y'\in Y} e^{-E(\textbf{x},y')}\big)$.



In machine learning, there is a deep relationship between the EBMs and discriminative models, which can be seen by connecting the Gibbs distribution in Equation (\ref{eq:gibbs_dist}) and the categorical distribution derived for a discriminative model.
A discriminative classifier is defined as a function for mapping the input $\textbf{x}$ to $C$ real-valued logits (i.e., for $C$ number of class labels): $f(\textbf{x}): \mathbb{R}^D \rightarrow \mathbb{R}^C$. In order to derive a categorical distribution over $C$  possible outputs, the softmax function is utilized:
\begin{equation}
\label{eq:categorical_dist}
    p(y|\textbf{x}) = \frac{e^{f_y(\textbf{x})}}{\sum_i^C e^{f_i(\textbf{x})}},
\end{equation}
where $f_y (\textbf{x})$ denotes the logit (probability) of the $y$th class label. 
Based on the inherent connection between the Gibbs and categorical distributions defined in (\ref{eq:gibbs_dist}) and (\ref{eq:categorical_dist}), the energy function for a given input $(\textbf{x},y)$ can be defined as $E(\textbf{x},y)=-f_y(\textbf{x})$. 
The free energy function $F(\textbf{x};f)$ can then be obtained by taking the negative log of the categorical distribution denominator as:
\begin{equation}
\label{eq:classifier_free_energy}
    F(\textbf{x};f) = -log \sum_i^C e^{f_i(\textbf{x})}.
\end{equation}

\subsection{Energy-Based Joint Inference}
\label{ssec:joint_inference}
Our goal is to detect the easy samples suitable for the Swift, which are indeed the ones with high likelihood in the density function. The energy-based density function for Swift is then defined as:
\begin{equation}
\label{eq:classifier_energy_density}
   p(\textbf{x}) = \frac{e^{-F(\textbf{x};f)}}{\int_{\textbf{x}} e^{-F(\textbf{x};f)}},
\end{equation}
where the denominator is the normalized densities, which can be intractable to compute or estimate. By taking the logarithm of both sides, we obtain:
\begin{equation}
\label{eq:classifier_log_density}
    log \big( p(\textbf{x}) \big) = -F(\textbf{x};f) - log(\int_{\textbf{x}} e^{-F(\textbf{x};f)}).
\end{equation}


The $log(\int_{\textbf{x}} e^{-F(\textbf{x};f)})$ term has no effect on the distribution of the overall energy values because it is constant for all $\textbf{x}$. As a result, $-F(\textbf{x};f)$, i.e., the negative free energy, has a linear alignment with the log likelihood function, which makes it a well-suited solution to the easy vs. hard detection problem in our framework. To this end, lower energy values indicate higher likelihood and represent easier (more fit) samples for the Swift model.

More precisely, for a threshold $\delta$ on the density function such that $p(\textbf{x}) < \delta$, then a threshold $t$ on the negative free energy can be calculated according to (\ref{eq:classifier_log_density}) as $-F(\textbf{x};f) < t = log (\delta \int_{\textbf{x}} e^{-F(\textbf{x};f)})$. In practice, for a given input, an energy function is applied to the outputs of the Swift model during inference time to calculate the energy score. Then, if the negative energy value is smaller than a threshold, the input is identified as a bad sample for the Swift, and is sent to the Super model.

Given the energy threshold $t$, the Swift classifier $f(\textbf{x})$, and the Super classifier defined as $g(\textbf{x}): \mathbb{R}^D \rightarrow \mathbb{R}^{C}$, the joint inference function $J(\textbf{x};f,g,t) \in [1,C]$ for a classification task with $C$ classes can then be expressed by: 
\begin{equation}
\label{eq:classifier_jr}
    J(\textbf{x};f,g,t) = \begin{cases} \mbox{$f(\textbf{x})$} & \mbox{if } - F(\textbf{x};f) \geq t \\ \mbox{$g(\textbf{x})$} & \mbox{otherwise.} \end{cases}
\end{equation}

\subsubsection{Encoder-Decoder Architectures}
\label{ssec:generative}

The proposed energy-based joint inference solution can be directly applied to the encoder-only models such as BERT that are designed for text classification tasks. To this end, the energy scores corresponding to the BERT-based Swift model are obtained using Equation (\ref{eq:classifier_free_energy}) and the joint inference is performed based on Equation \ref{eq:classifier_jr}.

On the other hand, for the encoder-decoder (auto-encoder) architectures such as T5, which are usually considered as generative models, some modifications are required. Encoder-decoder models are basically designed for sequence-to-sequence (e.g., text-to-text) problems such as translation or summarization. Although such models can also be employed for classification tasks, they still consider the task as a text generation (sequence-to-sequence) problem, where the target labels and the output predictions are treated as a sequence or a piece of text. 

In Section \ref{ssec:energy_definition}, it was discussed that there is an inherent connection between the discriminative classifiers and the EBMs. In order to benefit from this characteristic for encoder-decoder architectures, we consider adding an extra classification head (i.e., a single linear layer) to the Swift model. As encoders are commonly considered as better feature extractors for training a classifier rather than the decoders, we place the extra head after the Swift encoder. While freezing the pre-trained encoder model (denoted by $f^E$), the extra energy head (denoted by $h$) is trained as a regular classifier head with $C$ class labels. Note that the decoder is not required for training the head. The corresponding free energy function is then defined as follows: 
\begin{equation}
\label{eq:generator_free_energy}
    F(\textbf{x};f^E,h) = -log \sum_i^C e^{h_i\big(f^E(x)\big)},
\end{equation}
where $f^E(x)$ denotes the outputs of the encoder's last hidden state. These features are then fed to the extra head $h$ to obtain the logits for the $i$th class required for computing the energy scores. 

In this approach, as the decoder part of the Swift model is not required for calculating the energy scores, less computations are involved and the joint inference is performed more efficiently.

For text-to-text (or sequence-to-sequence) problems such as translation, the output is a sequence of $M$ word-pieces from a vocabulary/dictionary of size $N$. To still utilize the relationship of discriminative models and EBMs in designing and training the extra energy head,
we can treat the text-to-text models as $M$ multi-class classifiers. In this case, the number of class labels, i.e., $C$ in (\ref{eq:generator_free_energy}), is equal to $N$. The final energy score 
is then calculated as the average of $M$ energy values as follows: 
\begin{equation}
\resizebox{0.88\hsize}{!}{$
\label{eq:generator_free_energy2}
    \hspace{-8pt} F(\textbf{x};f^E,h) = - \frac{1}{M} \sum_m^M \big( log \sum_i^C e^{h_{m,i}\big(f^E(x)\big)} \big),
    \hspace{-4pt}
$}
\end{equation}
where $h_{m,i}(.)$ denotes the logits corresponding to the $m$th word in the sequence and $i$th class label.

Denote the Swift's decoder by $f^D$, the joint inference function, $J(\textbf{x};f,g,h,t)$, based on energy scores in either Equation (\ref{eq:generator_free_energy}) or (\ref{eq:generator_free_energy2}) is expressed as:
\begin{equation}
\label{eq:generator_jr}
    J = \begin{cases} \mbox{$f^D\big(f^E(x)\big)$} & \mbox{if } - F(\textbf{x};f^E,h) \geq t \\ \mbox{$g(\textbf{x})$} & \mbox{otherwise.} \end{cases}
\end{equation}

\subsection{Softmax and Entropy Mechanisms}
In addition to energy, softmax and entropy \citep{deebert} scores can also be used for analyzing the Swift model's performance in the routing mechanism. In this sub-section, we study the mathematical connection of them with the energy score and their potential to solve our problem.

\subsubsection{Softmax-Based Mechanism}
\label{sssec:softmax}

The softmax score for a classifier is expressed by:
\begin{equation}
\resizebox{0.86\hsize}{!}{$
\label{eq:softmax_score}
    \hspace{-8pt} \max_{y} p(y|\textbf{x}) = \max_{y} \frac{e^{f_y(\textbf{x})}}{\sum_i^C e^{f_i(\textbf{x})}} =  
    \frac{e^{f_{max}(\textbf{x})}}{\sum_i^C e^{f_i(\textbf{x})}}. \hspace{-4pt}
$}
\end{equation}

By taking the logarithm of both sides, 
we see the connection between the log of the softmax and the free energy score formulated in Equation (\ref{eq:classifier_free_energy}):
\begin{multline}
    log \max_{y} p(y|\textbf{x}) = 
    log (e^{f_{max}(\textbf{x})}) - log \sum_i^C e^{f_i(\textbf{x})}
    \\
    = f_{max}(\textbf{x}) + F(\textbf{x};f), \hspace{30pt}
\end{multline}
where all logits are shifted by their maximum  $f_{max}(x)$. Plugging in the energy term to (\ref{eq:classifier_log_density}) yields:
\begin{equation}
\label{eq:softmax_log_density}
     \begin{split}
        \hspace{-6pt} log \max_{y} p(y|\textbf{x}) = -log(p(\textbf{x})) + f_{max}(\textbf{x}) \\
        - log\big(\int_{\textbf{x}} e^{-F(\textbf{x};f)}\big). \hspace{15pt}
     \end{split}
\end{equation}

It is observed that for the samples with high likelihood of being in the Swift's distribution, the free energy goes lower, but the max logit tends to go higher. Due to this shifting, unlike the energy score, the softmax score is not well-aligned with the probability density $p(\textbf{x})$. 
As a result, the softmax score is less reliable for our routing module to analyze the performance of the Swift.


\subsubsection{Entropy-Based Mechanism}
\label{sssec:entropy}

The entropy score is a measure of randomness in the processed information, and is calculated as:
\begin{equation}
    H(\textbf{x};f)=-\sum_i^C f_i.log (f_i),
\end{equation}
where $f_i(\textbf{x})$ is the probability (logit) corresponding to the $i$th class label. Let $U$ be the internal energy, i.e., the expectation value of the energy function \citep{oh2020entropy}, defined by:
\begin{equation}
    U(\textbf{x};f) = \sum_i^C E(\textbf{x},i) f_i.
\end{equation}

According to \citet{oh2020entropy}, the entropy can be defined in terms of the internal and free energy functions as: $H(\textbf{x};f) = U(\textbf{x};f) - F(\textbf{x};f),$
where all logits are shifted by the internal energy $U$. 
Substituting the free energy from (\ref{eq:classifier_log_density}) yields:
\begin{equation}
\resizebox{0.85\hsize}{!}{$
    \hspace{-10pt} H(\textbf{x};f) = log(p(\textbf{x})) + U(\textbf{x};f) + log\big(\int_{\textbf{x}} e^{-F(\textbf{x};f)}\big),
    \hspace{-4pt} 
$}
\end{equation}
which shows, due to the shifting caused by internal energy, the entropy is not reliably aligned with the probability density $p(\textbf{x})$. Thus, it is a less suitable routing mechanism unlike the energy score. 

\section{Experimental Results}
\label{sec:results}

In this section, the performance of E-LANG on different architectures such as T5 and BERT; and benchmarks such as GLUE \citep{wang2019glue}, SuperGLUE \citep{wang2019superglue}, and WMT \citep{bojar2016findings} is evaluated and compared with the Super models and previous works. 

\begin{table*}[!tb]
\fontsize{7.5}{10}\selectfont
\begin{center}
\begin{tabular}[t]{p{0.5cm}lp{0.5cm}p{0.5cm}p{0.4cm}p{0.35cm}p{0.55cm}p{0.5cm}p{0.4cm}p{0.5cm}p{0.5cm}p{0.5cm}p{0.4cm}p{0.4cm}p{0.4cm}p{0.4cm}}
\toprule
\multicolumn{2}{c}{ } & 
\multicolumn{6}{c}{\textbf{GLUE}} & \multicolumn{7}{c}{\textbf{SuperGLUE}} & \multicolumn{1}{c}{\textbf{WMT}}\\
\\[-0.25cm]
\cmidrule(l{6pt}r{0pt}){3-8} 
\cmidrule(l{6pt}r{0pt}){9-15}
\cmidrule(l{6pt}r{0pt}){16-16} 
\\[-0.25cm]
 & & \textbf{MNLI} & \textbf{QNLI} & \textbf{SST2} & \textbf{RTE} & \textbf{MRPC} & \textbf{COLA} & \textbf{RTE} & \textbf{BoolQ} & \textbf{MRC} & \textbf{COPA} & \textbf{CB} & \textbf{WIC} & \textbf{WSC} & \textbf{EnRo}
\\
\\[-0.35cm]
\midrule
\hspace{-5pt} \multirow{2}{*}{\rotatebox[origin=c]{90}{\shortstack{\textbf{Swift}\\ \tiny(Large)}}} & 
\hspace{-10pt}
Time \tiny{(ms)} & {216} & {283} & {57} & {263} & {160} & {56} & {287} & {303} & {201} & {96} & {223} & {185} & {133} & 1609 \\
 & \hspace{-10pt} Accuracy \tiny{($\%$)} & 89.7 & 93.9 & 95.5 & 90.3 & 90.9 & 62.7 & 88.5 & 84.3 & 80.7 & 81.0 & 92.0 & 72.7 & 86.5 & 28.6 \\
\\[-0.25cm]
\midrule
\hspace{-5pt} \multirow{2}{*}{\rotatebox[origin=c]{90}{\shortstack{\textbf{Super}\\ \tiny(11B)}}} &  
\hspace{-10pt}
Time \tiny{(ms)} & {821} & {980} & {281} & {964} & {433} & {213} & 818 & 3205 & 1731 & 268 & 844 & 671 & 2211 & 3041 \\
& \hspace{-10pt} Accuracy \tiny{($\%$)} & 91.7 & 95.9 & 96.6 & 92.4 & 91.7 & 69.1 & 93.1 & \textbf{89.4} & 84.9 & \textbf{93.0} & 93.1 & 77.4 & 89.4 & 28.9 \\
\\[-0.2cm]
\midrule
\hspace{-5pt} \multirow{6}{*}{\textbf{\rotatebox[origin=c]{90}{E-LANG}}}  
& \hspace{-10pt} Accuracy \tiny{($\%$)} & \textbf{91.7} & \textbf{96.0} & \textbf{96.6} & \textbf{92.4} & \textbf{92.2} & \textbf{69.5} & \textbf{93.2} & 88.7 & \textbf{84.9} & 90.0 & \textbf{93.1} & \textbf{78.1} & \textbf{89.4} & \textbf{28.9} 
\\
 & \hspace{-10pt} FLOPs \tiny{($\times10^{11}$)} & {47.8} & {25.7} & {29.5} & {50.4} & {11.5} & {39.9} & {42.0} & {50.8} & {46.9} & {52.6} & 13.4 & 40.3 & 20.6 & 63.4 \\
&  \hspace{-10pt} Time \tiny{(ms)} & {582} & {495} & {132} & {716} & {190} & {147} & 671 & 1978 & 1022 & 222 & 302 & 447 & 545 & 2800 \\
& \hspace{-10pt} Swift Ratio \tiny{($\%$)} & 49 & 75 & 70 & 46 & 91 & 58 & 56 & 45 & 50 & 43 & 89 & 57 & 81 & 30 \\
 & \hspace{-10pt} Speed-up \tiny{(FLOPs)} & {1.8X} & {3.4X} & {2.9X} & {1.7X} & {7.6X} & {2.2X} & {2.1X} & {1.7X} & {1.9X} & {1.7X} & 6.5X & 2.2X & 4.2X & 1.4X \\
&  \hspace{-10pt} Speed-up \tiny{(time)} & {1.4X} & {2.0X} & {2.1X} & {1.4X} & {2.3X} & {1.5X} & 1.2X & 1.6X & 1.7X & 1.2X & 2.8X & 1.5X & 4.1X & 1.1X  \\
\bottomrule
\end{tabular}
\end{center}
\vspace{-7pt}
\caption{\label{tbl:results}\small Joint inference results with T5 architecture on GLUE and SuperGLUE development sets, and WMT's English-to-Romanian translation. The FLOPs for Super and Swift are respectively 87$\times10^{11}$ and 4.25$\times10^{11}$.}
\end{table*}

\begin{figure*}[tb!]
    \centering
    \begin{subfigure}{0.34\textwidth}
        \centering
        \includegraphics[width=0.99\linewidth]{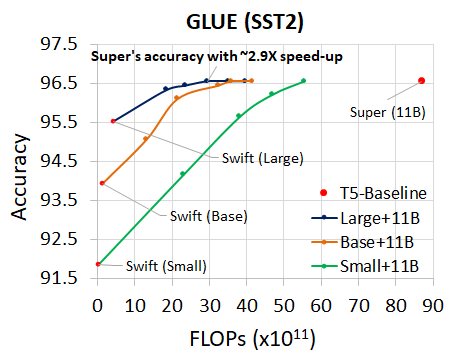}
    \end{subfigure}
    \hspace{-10pt}
    \begin{subfigure}{0.34\textwidth}
        \centering
        \includegraphics[width=0.99\linewidth]{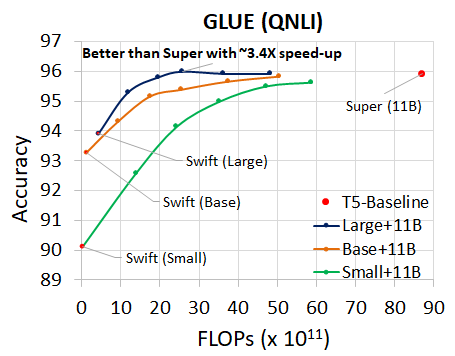}
    \end{subfigure}
    \hspace{-10pt}
    \begin{subfigure}{0.34\textwidth}
        \centering
        \includegraphics[width=0.99\linewidth]{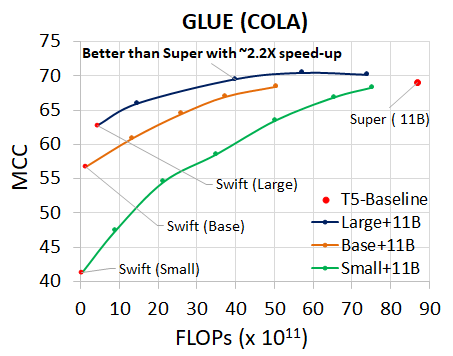}
    \end{subfigure}    
    \hspace{-10pt}
    \begin{subfigure}{0.34\textwidth}
        \centering
        \includegraphics[width=0.99\linewidth]{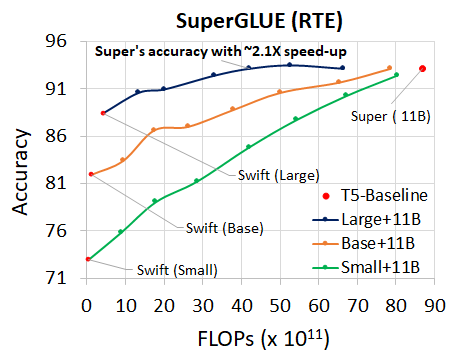}
    \end{subfigure}    
    \hspace{-10pt}
    \begin{subfigure}{0.34\textwidth}
        \centering
        \includegraphics[width=0.99\linewidth]{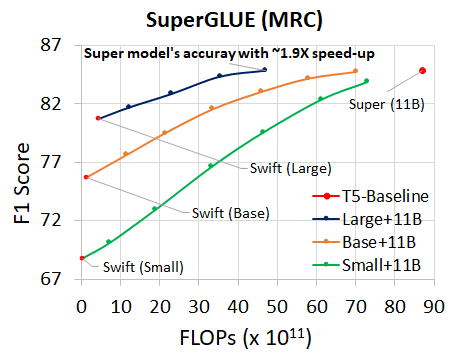}
    \end{subfigure}    
    \hspace{-10pt}
    \begin{subfigure}{0.34\textwidth}
        \centering
        \includegraphics[width=0.99\linewidth]{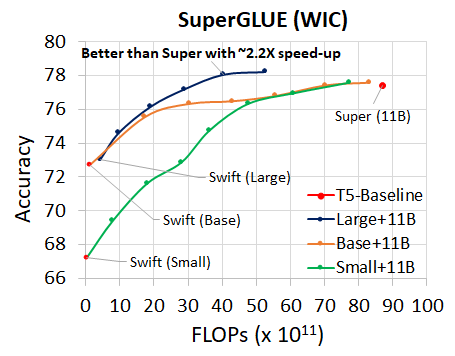}
    \end{subfigure} 
    \vspace{-7pt}
    \caption{\small Joint inference trade-off curves with T5 architecture on GLUE and SuperGLUE development sets. Each point is obtained with a different energy threshold.}
    \label{fig:glue_curves}
\end{figure*}

\subsection{T5-Based Joint Inference}
\label{ssec:t5_results}
In Table \ref{tbl:results}, the T5-based results on GLUE, SuperGLUE, and WMT benchmarks are reported. For all the tasks, we use T5-11B (with 87$\times10^{11}$ FLOPs) and T5-large (with 4.25$\times10^{11}$ FLOPs) as our Super and Swift models, respectively. The average GPU-based running time and accuracy of both models compared with E-LANG are also summarized in the table. Note that the T5 models used in this experiment have been separately fine-tuned on each of the downstream tasks given in Table \ref{tbl:results}. The extra energy head for each of these tasks was also separately trained and used based on the task-specific number of classes, i.e., $C$ in Equation (\ref{eq:generator_free_energy}). 

The total FLOPs for our method is measured as a weighted average of the Super and Swift FLOPs based on their usage frequency as:
\begin{multline}
FLOPs=\frac{1}{N_{sw}+N_{su}}\big(N_{sw}.(F^E_{sw}+F_{h}+F^D_{sw})
\\
+N_{su}.(F^E_{sw}+F_{h}+F_{su})\big), 
\end{multline}
where $N_{su}$ and $N_{sw}$ are respectively the number of samples processed by the Super (with $F_{su}$ FLOPs) and Swift (with $F^E_{sw}$, $F^D_{sw}$, and $F_h$ FLOPs for the encoder, decoder, and energy head). Note that $F_h$ is equal to $\approx$0.00001$\times10^{11}$, which has a very insignificant overhead in our framework.

As presented in Table \ref{tbl:results}, E-LANG can reach the Super model's accuracy on all GLUE tasks with an average 3.3X in FLOPs and 1.8X in running time speed-ups. For some tasks such as QNLI, MRPC, and COLA, we even outperform the Super model, which leads to a higher average accuracy of 89.7\% than the Super model with 89.5\% on GLUE. For the SuperGLUE benchmark, with an average FLOPs and running time speed-up of 2.9X and 2.0X, our method achieves the same accuracy as the Super model on MRC and CB; and better accuracy on RTE and WIC. On BoolQ and COPA, although 99\% and 97\% of the Super's accuracy are respectively obtained, it is with 1.7X and 1.4X less FLOPs and latency, on average.

In order to analyze the generality of E-LANG to other NLP problems rather than text classification (Section \ref{ssec:generative}), we also apply our method to two text-to-text tasks including SuperGLUE's WSC and WMT's English-to-Romanian (EnRo) translation. As given in the table, our method achieves the Super model's accuracy on both WSC and EnRo with 4.2X and 1.4X less FLOPs, respectively.


%


 

%

Figure \ref{fig:glue_curves} illustrates the accuracy vs. FLOPs trade-off curves
for some tasks in GLUE and SuperGLUE benchmarks. The curves related to all tasks are given in the supplementary materials. The trade-off points on the curves are dynamically achieved at the inference time by selecting different thresholds, i.e., $t$ in Equations (\ref{eq:classifier_jr}) and (\ref{eq:generator_jr}). Larger values for $t$ will result in routing more input data to the Super model, which consequently provides more accurate, but slower inference. As the Swift is able to make accurate predictions for the majority of input data, the dynamic inference with a small enough $t$ can reach the Super model's accuracy but with a much lower computational cost and latency.

\begin{figure*}[tb!]
    \centering
    \begin{subfigure}{0.34\textwidth}
        \centering
        \includegraphics[width=0.99\linewidth]{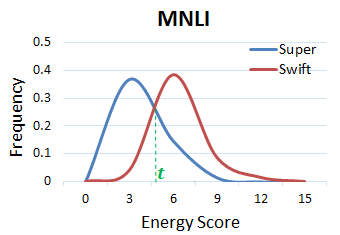}
    \end{subfigure}
    \hspace{-10pt}
    \begin{subfigure}{0.34\textwidth}
        \centering
        \includegraphics[width=0.99\linewidth]{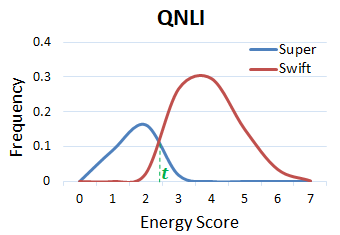}
    \end{subfigure}
    \hspace{-10pt}
    \begin{subfigure}{0.34\textwidth}
        \centering
        \includegraphics[width=0.99\linewidth]{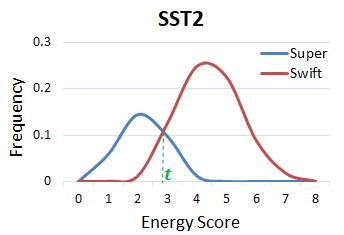}
    \end{subfigure}    
    \hspace{-10pt}
    \begin{subfigure}{0.34\textwidth}
        \centering
        \includegraphics[width=0.99\linewidth]{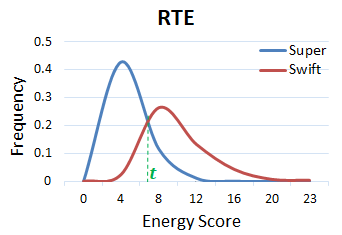}
    \end{subfigure}
    \hspace{-10pt}
    \begin{subfigure}{0.34\textwidth}
        \centering
        \includegraphics[width=0.99\linewidth]{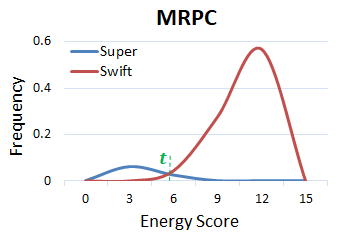}
    \end{subfigure}
    \hspace{-10pt}
    \begin{subfigure}{0.34\textwidth}
        \centering
        \includegraphics[width=0.99\linewidth]{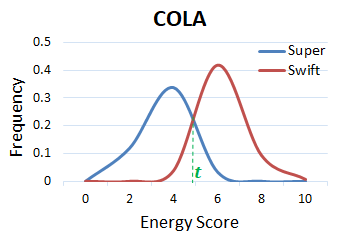}
    \end{subfigure}  
    \vspace{-7pt}
    \caption{\small Energy score distribution for GLUE tasks. $t$ shows the optimal threshold.}
    \label{fig:energy_distribution}
\end{figure*}

Figure \ref{fig:energy_distribution} illustrates the distribution of the energy scores across the input samples in GLUE tasks. For each task, the distributions of the samples processed by the Super and the Swift models are plotted. As shown, the samples routed to the Super model tend to have lower energy scores that are indeed considered as out-of-distribution samples for the Swift. On the other hand, in overall, higher scores are observed for the Swift distribution, that is for the samples handled by the Swift only. For some tasks such as MRPC and QNLI, the Swift is shown to be highly capable of handling the majority of the input samples. This is also supported by the results in Table \ref{tbl:results} and Figure \ref{fig:glue_curves}, where 91\% (for MRPC) and 75\% (for QNLI) of the samples are accurately processed by the Swift. In contrast, for other datasets including RTE and MNLI with Swift ratio of less than 50\%, most of the samples are hard for the Swift, which are transferred to the Super model. Based on our experiments, the most optimal results for our joint inference framework is achieved when the crossing point of the two distributions (highlighted in green in the figures) is chosen as the threshold $t$ in Equation (\ref{eq:generator_jr}).

\subsubsection{Ablation Studies}
\label{sssec:t5_ablation}

In Sections \ref{sssec:softmax} and \ref{sssec:entropy}, the possibility of using softmax and entropy scores instead of energy score was theoretically analyzed. To support that analysis and also experimentally evaluate the performance of different routing mechanisms, an ablation study on GLUE is performed, which is presented in Table \ref{tbl:ablation_studies}. In this study, we report the joint inference results based on softmax, entropy, and random scores (i.e., randomly distributing the samples between Super and Swift). Our experiments show that, compared to the random score, softmax and entropy can result in satisfactory performance on routing the samples. However, as also discussed in Sections \ref{sssec:softmax} and \ref{sssec:entropy}, the energy score is still a better mechanism with about 14\% less FLOPs. Another potential mechanism is the perplexity \citep{chen1998evaluation}, but since it provides the same information as entropy, we did not add any extra experiment on it.

The results with the usage of different Swift models including T5-small (with 0.33$\times10^{11}$ FLOPs) and T5-base (with 1.24$\times10^{11}$ FLOPs) are also given in Table \ref{tbl:ablation_studies}. Using these models as Swifts can lead to good performance on some tasks, but not all of them. For example, on SST2, the joint inference with T5-small and T5-base Swifts can respectively reach the Super's accuracy with 1.9X and 2.X less computations. In general, although these models are smaller and require less FLOPs, our results in Table \ref{tbl:ablation_studies} indicate that they perform worse than T5-large in our joint inference structure. In Figure \ref{fig:glue_curves}, the trade-off curves for different Swift models are shown for GLUE and SuperGLUE.

Moreover, to show the effectiveness of the extra energy head for the Swift encoder, the E-LANG results based on last linear layer of the Swift decoder is also given and compared in Table \ref{tbl:ablation_studies}. As reported, the E-LANG empowered by the energy head on the Swift encoder outperforms the case with the decoder's head in both FLOPs (36.8\% less) and accuracy (0.7\% better). As explained in Section \ref{ssec:generative}, this shows the deep connection between the encoder's features, discriminative models, and the proposed routing mechanism via the energy head.


\begin{table*}[!tb]
\fontsize{7.5}{10}\selectfont
\begin{center}
\begin{tabular}[t]{p{2.4cm}p{1.4cm}p{1.4cm}p{1.4cm}p{1.4cm}p{1.4cm}p{1.4cm}|p{1.4cm}}
\toprule
& \textbf{MNLI} & \textbf{QNLI} & \textbf{SST2} & \textbf{RTE} & \textbf{MRPC} & \textbf{COLA} & \textbf{Average}  
\\
\\[-0.35cm]
\midrule
\hspace{-5pt} \textbf{Super} \tiny(11B) & 87.0 / 91.7 & 87.0 / 95.9 & 87.0 / 96.6 & 87.0 / 92.4 & 87.0 / 91.7 & 87.0 / 69.1 & {87.0 / 89.5}
 \\
\midrule
\hspace{-5pt} \textbf{Random} \tiny(Encoder) & 78.5 / 91.5 & 61.9 / 95.3 & 58.7 / 96.3 & 60.2 / 91.2 & 47.5 / 91.9 & 61.6 / 67.2 & {61.4 / 88.9}
 \\
\hspace{-5pt} \textbf{Softmax} \tiny(Encoder) & 57.7 / 91.6 & 36.5 / 95.9 & 34.6 / 96.5 & 52.0 / 92.3 & 13.8 / 92.1 & 45.7 / 69.3 & {40.1 / 89.6}
 \\
\hspace{-5pt} \textbf{Entropy} \tiny(Encoder) & 55.7 / 91.6 & 27.1 / 96.0 & 40.2 / 96.5 & 50.7 / 92.0 & 23.0 / 92.2 & 48.1 / 69.3 & {40.8 / 89.6}
 \\
\hspace{-5pt} \textbf{Energy} \tiny(Swift$_{small}$) & 71.3 / 91.0 & 58.8 / 95.6 & 47.0 / 96.6 & 71.2 / 88.5 & 55.0 / 91.4 & 75.3 / 68.3 & {63.1 / 88.5} \\
\hspace{-5pt} \textbf{Energy} \tiny(Swift$_{base}$) & 54.5 / 91.5 & 50.5 / 95.8 & 35.9 / 96.6 & 55.8 / 90.6 & 44.0 / 91.9 & 50.6 / 68.4 & {48.5 / 89.1} \\
\hspace{-5pt} \textbf{Energy} \tiny(Decoder) & 57.9 / 90.6 & 68.1 / 95.5 & 75.8 / 96.3 & 60.5 / 91.5 & 20.2 / 90.9 & 45.1 / 69.3 & {54.6 / 89.0}
 \\
\hspace{-5pt} \textbf{Energy} \tiny(Encoder) & \textbf{47.8 / 91.7} &\textbf{25.7 / 96.0} & \textbf{32.0 / 96.6} & \textbf{50.4 / 92.4} & \textbf{11.5 / 92.2} & \textbf{39.9 / 69.5} & \textbf{34.5 / 89.7}
 \\
\bottomrule
\end{tabular}
\end{center}
\vspace{-7pt}
\caption{\label{tbl:ablation_studies}\small Ablation study on different T5-based scenarios. Each cell shows FLOPs/Accuracy.}
\end{table*}

\begin{table}[tb!]
\fontsize{7.5}{10}\selectfont
\begin{center}
\begin{tabular}[t]{p{1.7cm}p{1.3cm}p{1.3cm}p{1.3cm}}
\toprule
& \textbf{QNLI} 
& \textbf{SST2} & \textbf{COLA} 
\\
\\[-0.35cm]
\midrule
\hspace{-5pt} {\textbf{Super} \tiny(11B)} 
& 87.0 / 95.9 
& 87.0 / 96.6 & 87.0 / 69.1
 \\
\hspace{-5pt} {\textbf{Swift} \tiny(Large)} 
& 4.25 / 93.9 
& 4.25 / 95.5 & 4.25 / 62.7
 \\
\hspace{0pt} \textbf{+ Distillation} 
& 4.25 / 95.0 
& 4.25 / 95.7 & 4.25 / 63.3
 \\
\hspace{-5pt} \textbf{Ours} 
& {25.7 / 96.0} 
& {29.5 / 96.6} & 39.9 / 69.5
 \\ 
\hspace{0pt} \textbf{+ Distillation} 
& \textbf{18.2 / 96.0} 
& \textbf{15.2 / 96.6} & \textbf{34.2 / 69.5}
 \\
\bottomrule
\end{tabular}
\end{center}
\vspace{-7pt}
\caption{\label{tbl:t5_distill}\small Distillation-based results with T5 in terms of FLOPs/Accuracy.}
\end{table}

We observed that E-LANG can achieve a high performance even when applied to individually pre-trained Super and Swift models. But, more improvement can still be obtained by performing KD from the Super to the Swift model, especially at the fine-tuning process for downstream tasks. To study this, we apply the KD technique in \citep{distilbert} to the Super and Swift models for some GLUE tasks. As summarized in Table \ref{tbl:t5_distill}, the Super model's accuracy for QNLI, SST2, and COLA is respectively attained by the distillation-based E-LANG with 29.2\%, 48.5\%, and 14.3\% less FLOPs than E-LANG (without distillation). The results show the effectiveness of E-LANG along with other compression techniques such as distillation. The trade-off curves for this experiment will be provided in the supplementary materials.

\subsection{BERT-Based Joint Inference}
\label{ssec:bert_results}

In this section, the proposed energy-based joint inference method is applied to the BERT architecture \citep{bert} and compared with BERT-based SOTA in both fixed-size and dynamic inference. The majority of the previous methods employ knowledge distillation and data augmentation techniques for training their student models. For a fair comparison, we follow the same practice and use the transformer distillation and augmentation strategies in TinyBERT \citep{tinybert} to train and prepare our Swift model (i.e., BERT$_{Tiny}$ with $1.2\times10^{9}$ FLOPs). Moreover, similar to the other works, we use BERT$_{Base}$ (with $21.8\times10^{9}$ FLOPs) as our Super (i.e., teacher) model.

In Table \ref{tbl:bert_results}, the comparison results with the baseline BERT$_{Base}$ and SOTA on GLUE benchmark are presented in terms of accuracy, FLOPs, and latency. Compared to the Super model, E-LANG delivers better accuracy on SST2 and RTE with 3.5X and 2.0X FLOPs speed-up; and the same accuracy on QNLI, MRPC, and QQP with 2.4X, 2.7X, and 7.0X FLOPs speed-up, respectively. On MNLI and COLA, 99.8\% and 97.3\% of the Super model's accuracy are achieved, but with an average FLOPs speed-up of 2.3X. On average, E-LANG outperforms the Super model with 0.1\% higher accuracy, 3.2X less FLOPs, and 1.6X less latency.

Compared with SOTA, our method achieves the best performance on all GLUE tasks, except MRPC for which SqueezeBERT outperforms all due to having a more accurate teacher \citep{squeezebert}. There are some works such as ELECTRA \citep{electra} and MobileBERT \citep{mobilebert} that require less FLOPs than our method, but they only reach 95\% of the baseline's accuracy. 
Compared to other methods, GhostBERT \citep{ghostbert} and DynaBERT \citep{dynabert} give the closest performance to the baseline and even the same as ours on some tasks such as QNLI. However, on average, they still need about 30\% more FLOPs on GLUE compared to E-LANG. 

The E-LANG accuracy vs. FLOPs trade-off curves compared to SOTA on some of GLUE tasks are shown in Figure \ref{fig:bert_glue_curves}. The trade-off curves for all the tasks are reported in the supplementary materials. Among the SOTA methods presented in Table \ref{tbl:bert_results} and Figure \ref{fig:bert_glue_curves}, only DeeBERT \citep{deebert}, Length-Adaptive \citep{lengthadaptive}, and DynaBERT \citep{dynabert} are in the category of dynamic inference, where a single model can operate at different trade-off points between accuracy and computational cost. The other approaches propose fixed-size smaller versions of BERT$_{Base}$, which require re-training for every trade-off point.



\begin{table*}[tb!]
\fontsize{7.5}{10}\selectfont
\begin{center}
\begin{tabular}[t]{p{0.3cm}lp{1.5cm}p{0.6cm}p{0.6cm}p{0.6cm}p{0.7cm}p{0.7cm}p{0.7cm}|p{0.7cm}p{0.9cm}p{0.8cm}}
\toprule
& & \textbf{MNLI}~\tiny{(m/mm)} & \textbf{QNLI} & \textbf{SST2} & \textbf{RTE} & \textbf{MRPC} & \textbf{COLA} & \textbf{QQP} & \textbf{Avg.} & \textbf{FLOPs}~\tiny{(G)} & \textbf{Time}~\tiny{(ms)} 
\\
\\[-0.35cm]
\midrule
\hspace{-5pt} \multirow{14}{*}{\textbf{\rotatebox[origin=c]{90}{Previous works}}} 
& \hspace{-10pt} BERT$_{Tiny}$ \tiny{(Swift)}  & 82.8 / 82.9 & 87.9 & 92.6 & 65.7 & 85.8 & 49.7 & 90.5 & 78.5 & 1.2 & 7
 \\
& \hspace{-10pt} BERT$_{Base}$ \tiny{(Super)} & \textbf{84.9 / 85.5} & 92.2 & 93.5 & 71.1 & 87.3 & \textbf{60.3} & 91.5 & 83.3 & 21.8 & 20 
 \\
 \midrule
& \hspace{-10pt} DistillBERT & 82.2 / - & 89.2 & 92.7 & 59.9 & 87.5 & 51.3 & 88.5 & 78.8 & 11.3  & -
 \\
& \hspace{-10pt} ELECTRA & 78.9 / - & 87.9 & 88.3 & 68.5 & 84.4 & 56.8 & 88.3 & 79.0 & 3.7 & -
 \\
& \hspace{-10pt} DeeBERT & 83.9 / 82.9 & 90.9 & 93.4 & 69.5 & - & - & - & - & - & 17
 \\
& \hspace{-10pt} MobileBERT  & 84.3 / - & 91.5 & 92.5 & 70.4 & 87.0 & 51.1 & - & 79.5 & 5.7 & -
 \\
& \hspace{-10pt} SqueezeBERT & 82.5 / 82.9 & 90.9 & 92.2 & 71.8 & \textbf{89.8} & 53.7 & 89.5 & 81.7 & 7.4 & -
 \\
& \hspace{-10pt} Len-Adaptive & 84.4 / - & - & 93.1 & - & - & - & - & - & 8.8 & -
 \\
& \hspace{-10pt} TinyBERT & 84.5 / 84.5 & 91.8 & 93.0 & 69.3 & 87.2 & 54.0 & 91.0 & 81.9 & 11.3 & 10 
 \\
& \hspace{-10pt} ELM & 84.2 / - & 90.8 & 92.7 & 72.2 & 89.0 & 54.2 & 91.1 & 82.0 & 10.9 & -
 \\
& \hspace{-10pt} GhostBERT & 84.7 / - & 92.2 & 92.9 & 72.2 & 87.3 & 58.1 & 91.2 & 82.7 & 11.3 & - 
\\
& \hspace{-10pt} DynaBERT & 84.7 / 85.2 & 92.2 & 93.3 & 73.0 & 84.8 & 58.4 & 91.3 & 82.9 & 10.9 & 16 
 \\
\midrule
\midrule
\hspace{-5pt} 
\multirow{6}{*}{\textbf{\rotatebox[origin=c]{90}{E-LANG}}} 
& \hspace{-10pt} \textbf{Accuracy} \tiny{($\%$)} & {84.7 / 85.4} & \textbf{92.2} & \textbf{93.7} & \textbf{73.3} & {87.3} & {58.7} & \textbf{91.5} & \textbf{83.4} & - & - \\
& \hspace{-10pt} \textbf{FLOPs} \tiny{(G)} & \textbf{9.1} & \textbf{9.2} & \textbf{6.3} & \textbf{10.8} & {8.2} & \textbf{9.9} & \textbf{3.1} & 8.1 & - & - \\
& \hspace{-10pt} \textbf{Time} \tiny{(ms)} & 14 & 14 & 11 & 16 & 13 & 15 & 9 & 13 & - & - \\
& \hspace{-10pt} \textbf{Swift Ratio} \tiny{(\%)} & 64 & 63 & 77 & 56 & 68 & 60 & 91 & 68 & - & - \\
& \hspace{-10pt} \textbf{Speed-up} \tiny{(FLOPs)} & 2.4X & 2.4X & 3.5X & 2.0X & 2.7X & 2.2X & 7.0X & 3.2X & - & -  \\
& \hspace{-10pt} \textbf{Speed-up} \tiny{(time)} & 1.4X & 1.4X & 1.8X & 1.3X & 1.5X & 1.3X & 2.2X & 1.6X & - & - \\
\bottomrule
\end{tabular}
\end{center}
\vspace{-7pt}
\caption{\label{tbl:bert_results}\small Joint inference results with BERT architecture on GLUE development set compared with SOTA.}
\end{table*}

\begin{figure*}[tb!]
    \centering
    \begin{subfigure}{0.34\textwidth}
        \centering
        \includegraphics[width=0.99\linewidth]{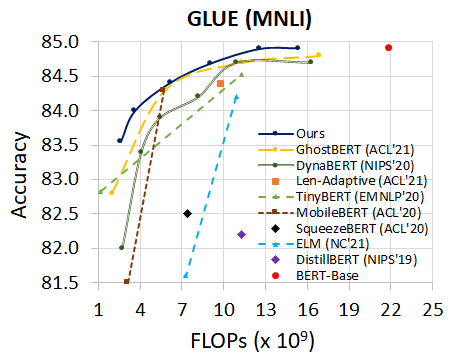}
    \end{subfigure}
    \hspace{-10pt}
    \begin{subfigure}{0.34\textwidth}
        \centering
        \includegraphics[width=0.99\linewidth]{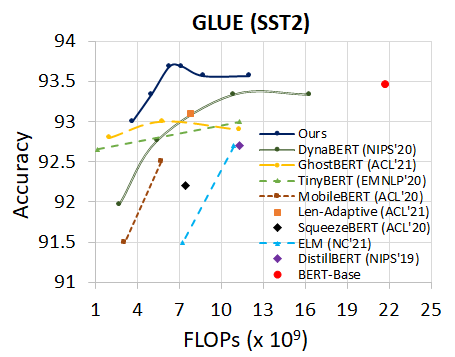}
    \end{subfigure}
    \hspace{-10pt}
    \begin{subfigure}{0.34\textwidth}
        \centering
        \includegraphics[width=0.99\linewidth]{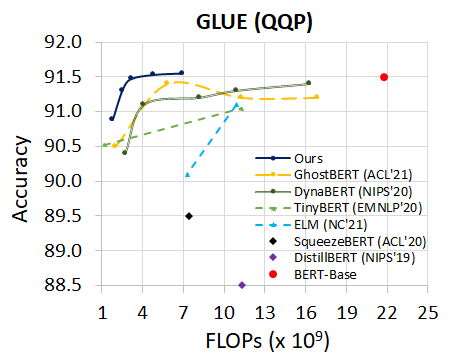}
    \end{subfigure}    
    \vspace{-7pt}
    \caption{\small Joint inference trade-off curves with BERT on GLUE development set compared with SOTA.}
    \label{fig:bert_glue_curves}
\end{figure*}

To investigate the orthogonality of E-LANG with others, we integrate our energy-based joint inference strategy with DynaBERT that is SOTA in BERT-based adaptive inference. In other words, we analyze whether E-LANG can be added on top of other efficient methods to benefit both from their designs and our approach. In this experiment, the DynaBERT configurations with the highest accuracy (i.e., width=0.75 \& depth=1.0) and the lowest FLOPs (i.e., width=0.5 \& depth=0.25) are respectively employed as the Super and Swift models in our framework. The corresponding joint inference results on MNLI, SST2, and QQP are reported in Table \ref{tbl:ortha_dynabert}. As observed, we accomplish the DynaBERT Super's accuracy for MNLI and SST2 with 1.7X and 3.1X less FLOPs. For QQP, our method combined with DynaBERT even outperforms DynaBERT by 0.1\% with 2.6X FLOPs speed-up.

\begin{table}[tb!]
\fontsize{7.5}{10}\selectfont
\begin{center}
\begin{tabular}[t]{p{2.0cm}p{1.3cm}p{1.3cm}p{1.3cm}}
\toprule
& \textbf{MNLI} & \textbf{SST2} & \textbf{QQP} 
\\
\\[-0.35cm]
\midrule
\hspace{-5pt} {\textbf{DynaBERT} \tiny(Swift) \tiny(w=0.5, d=0.25)} & 2.7 / 82.0 & 2.7 / 91.9 & 2.7 / 90.4
 \\[0.45cm]
\hspace{-5pt} {\textbf{DynaBERT} \tiny(Super) \tiny(w=0.75, d=1.0)} & 16.3 / 84.7 & 16.3 / 93.3 & 16.3 / 91.4
 \\[0.45cm]
\hspace{-3pt} \textbf{Ours+DynaBERT} & \textbf{9.4 / 84.7} & \textbf{5.2 / 93.3} & \textbf{6.2 / 91.5}
 \\
\bottomrule
\end{tabular}
\end{center}
\vspace{-7pt}
\caption{\label{tbl:ortha_dynabert}\small Orthogonality of E-LANG (ours) with DynaBERT in terms of FLOPs/Accuracy.}
\end{table}

\section{Conclusion}
\label{sec:conclusion}
In this paper, we introduced E-LANG, an energy-based joint inference approach, which integrates Super and Swift language models for achieving efficient inference without sacrificing the accuracy. Our method can work with both encoder-only (e.g., BERT) and encoder-decoder (e.g., T5) architectures, and is also applicable for text classification and sequence-to-sequence problems. The proposed joint inference strategy was theoretically and experimentally analyzed with an extensive set of experiments and ablation studies. Our results showed that E-LANG outperforms SOTA in both fixed-size and dynamic inference over different benchmarks such as GLUE and SuperGLUE. One future direction to this work is to apply E-LANG to multiple Super and Swift models with different sizes.


\bibliographystyle{acl_natbib}
\bibliography{ref.bib}

\clearpage
\appendix

\section{Supplementary Materials}
\label{sec:supp_mat}

This section contains the supplementary materials.

\subsection{Code and Demo}


We shared our code to make it easy to reproduce our BERT-based results. In addition to the code, we included a video demo that contains a demonstration of T5-based E-LANG. The BERT-based E-LANG source-code with the detailed running instructions and the T5-based E-LANG demo are available  \href{https://developer.huaweicloud.com/develop/aigallery/notebook/detail?id=64199726-9aaf-4905-8f6f-4cae290df874}{here\footnote{\href{https://developer.huaweicloud.com/develop/aigallery/notebook/detail?id=64199726-9aaf-4905-8f6f-4cae290df874}{https://developer.huaweicloud.com/develop/aigallery/notebook/detail?id=64199726-9aaf-4905-8f6f-4cae290df874}}}.

\begin{figure*}[!b]
    \centering
    \includegraphics[width=0.99\linewidth]{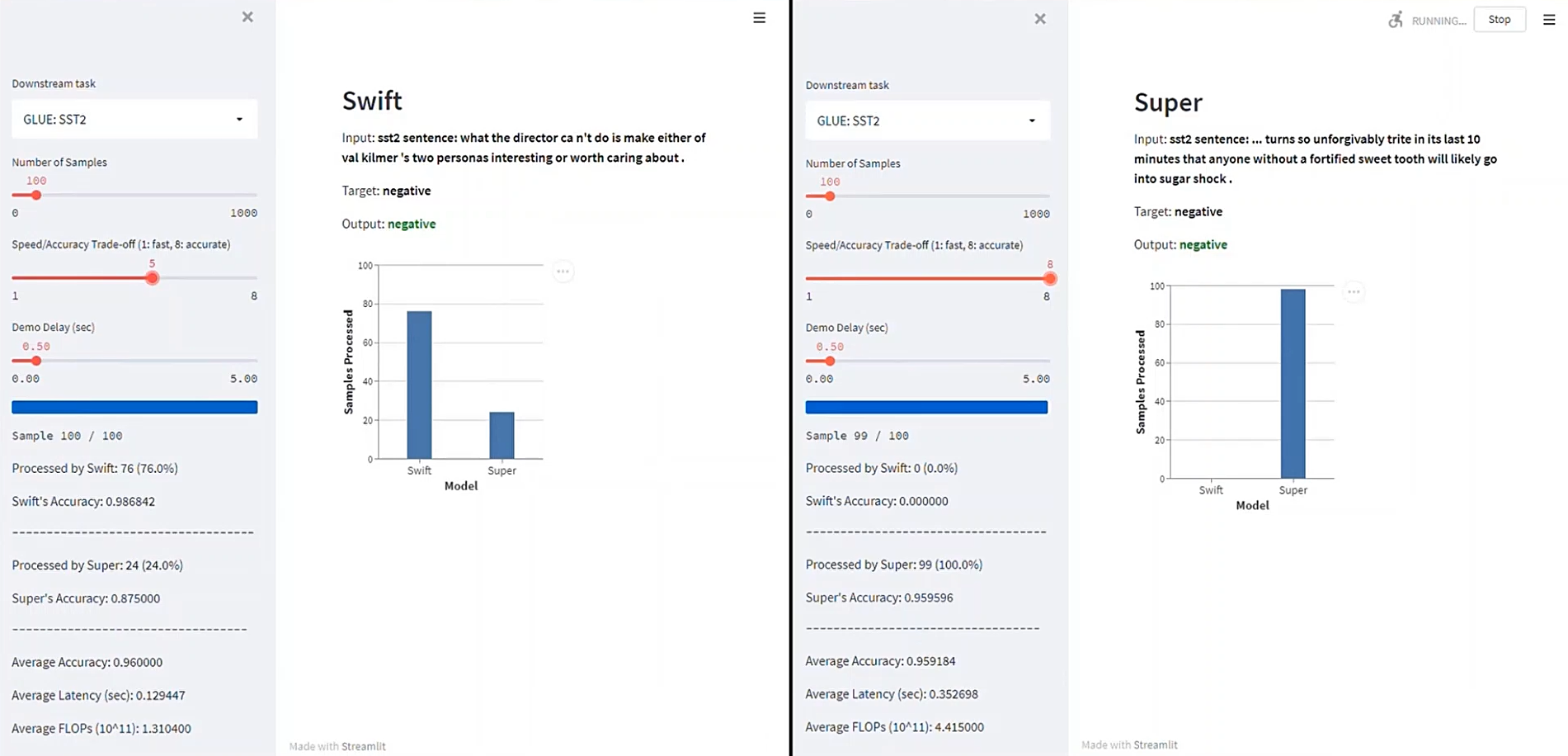}
    \caption{A demo application to show-case the adaptive inference with the proposed method.}
    \label{supp:demo_screenshot}
\end{figure*}

Please note that the demo is based on screen recording of a web application we built to show the use-cases of our method in real-world scenarios. Figure \ref{supp:demo_screenshot} shows a screenshot of the demo application. 


\subsection{Additional Results and Visualizations}

\begin{figure*}[b!]
    \centering
    \begin{subfigure}{0.50\textwidth}
        \centering
        \includegraphics[width=0.99\linewidth]{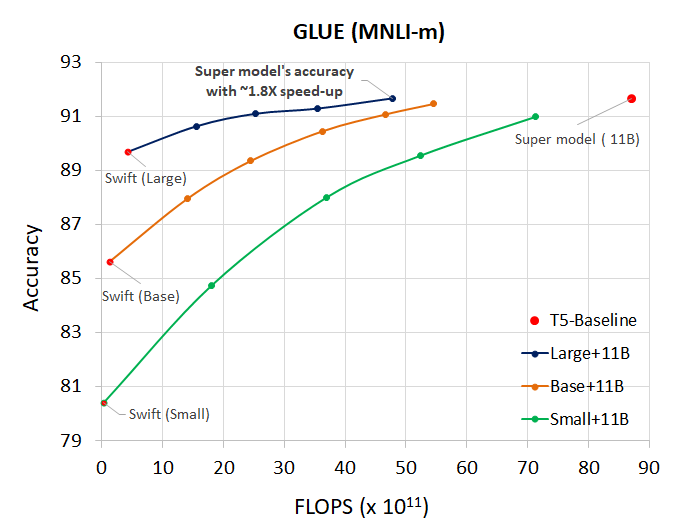}
    \end{subfigure}
    \hspace{-10pt}
    \begin{subfigure}{0.50\textwidth}
        \centering
        \includegraphics[width=0.99\linewidth]{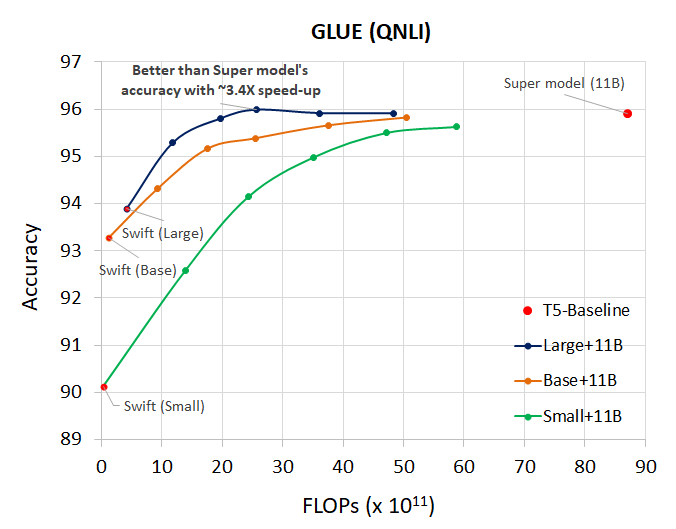}
    \end{subfigure}
    \hspace{-10pt}
    \begin{subfigure}{0.50\textwidth}
        \centering
        \includegraphics[width=0.99\linewidth]{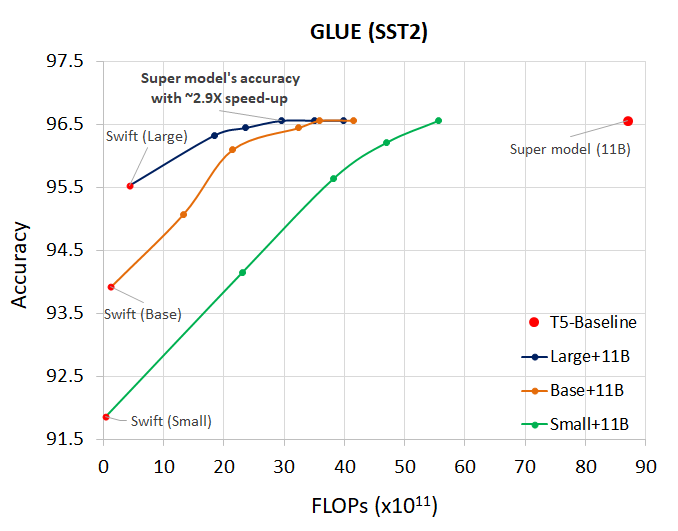}
    \end{subfigure}    
    \hspace{-10pt}
    \begin{subfigure}{0.50\textwidth}
        \centering
        \includegraphics[width=0.99\linewidth]{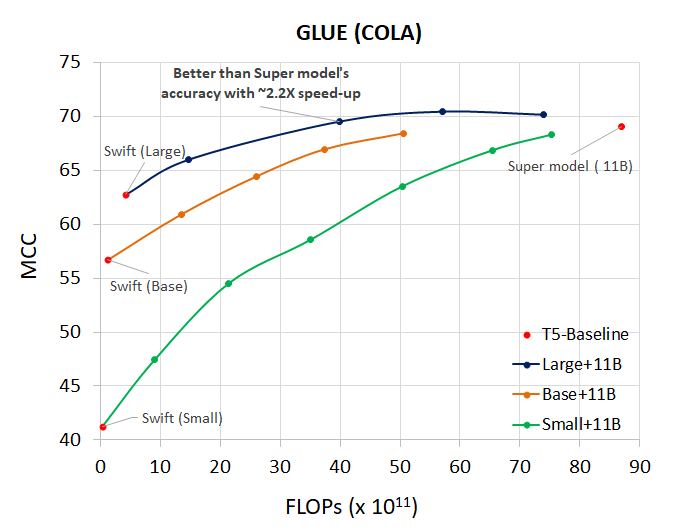}
    \end{subfigure}    
    \hspace{-10pt}
    \begin{subfigure}{0.50\textwidth}
        \centering
        \includegraphics[width=0.99\linewidth]{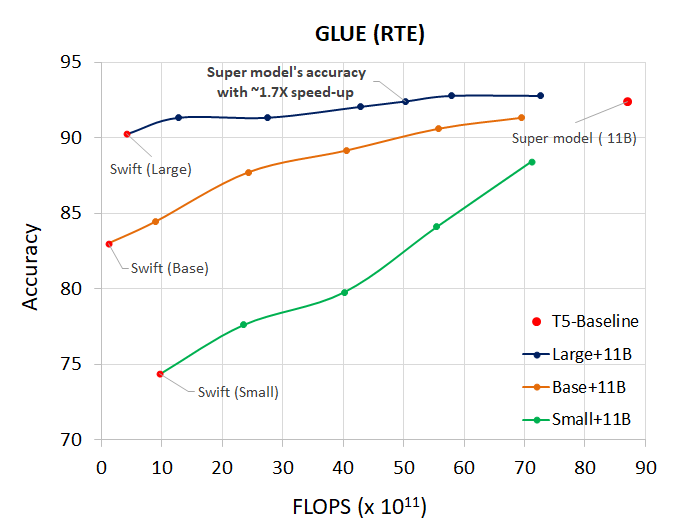}
    \end{subfigure}    
    \hspace{-10pt}
    \begin{subfigure}{0.50\textwidth}
        \centering
        \includegraphics[width=0.99\linewidth]{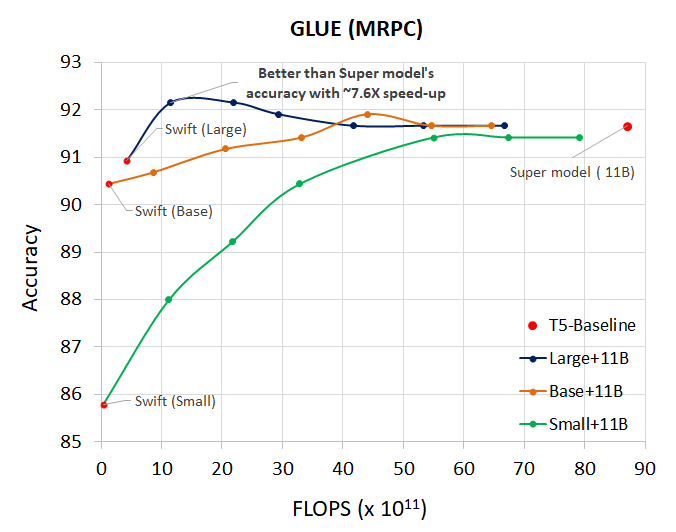}
    \end{subfigure}    
    \caption{\small Trade-off curves with T5 backbone on GLUE tasks.}
    \label{fig:glue_curves_appendix}
\end{figure*}

\begin{figure*}[b!]
    \centering
    \begin{subfigure}{0.49\textwidth}
        \centering
        \includegraphics[width=0.99\linewidth]{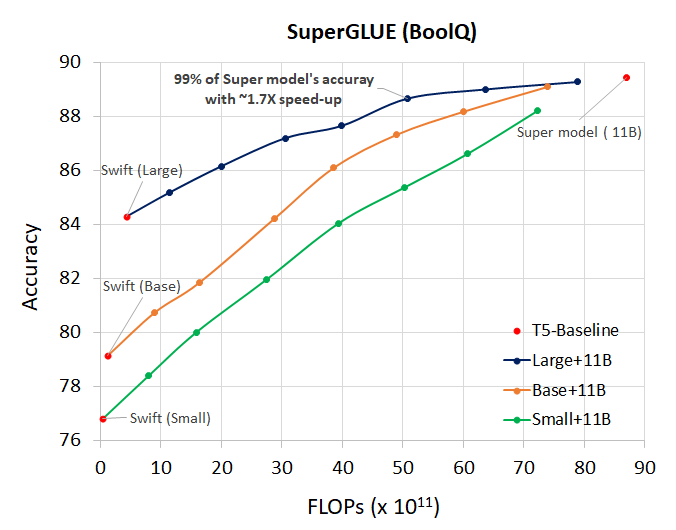}
    \end{subfigure}
    \hspace{-10pt}
    \begin{subfigure}{0.49\textwidth}
        \centering
        \includegraphics[width=0.99\linewidth]{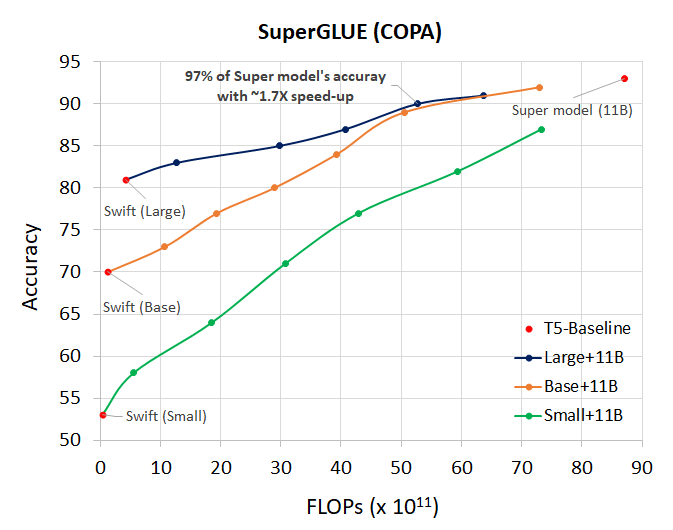}
    \end{subfigure}
    \hspace{-10pt}
    \begin{subfigure}{0.49\textwidth}
        \centering
        \includegraphics[width=0.99\linewidth]{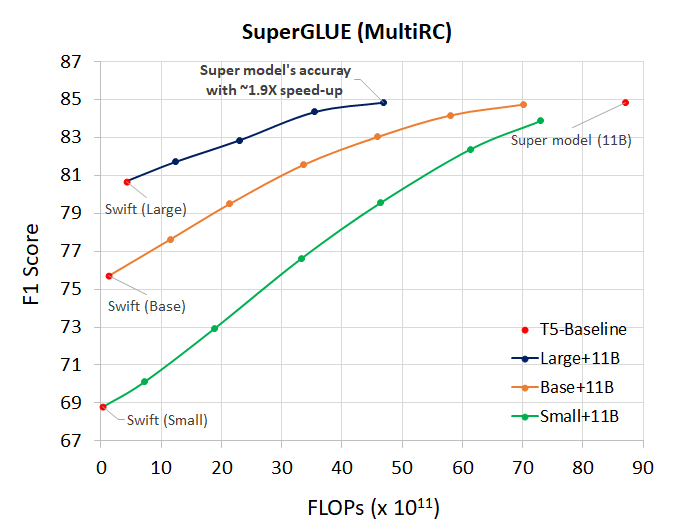}
    \end{subfigure}    
    \hspace{-10pt}
    \begin{subfigure}{0.49\textwidth}
        \centering
        \includegraphics[width=0.99\linewidth]{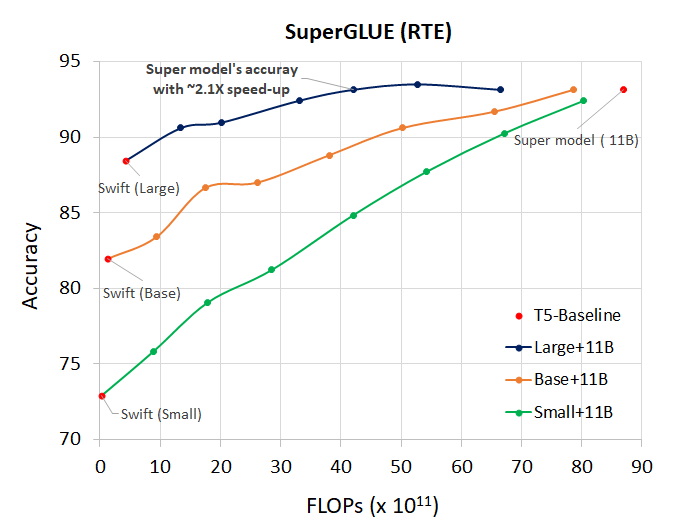}
    \end{subfigure}    
    \hspace{-10pt}
    \begin{subfigure}{0.49\textwidth}
        \centering
        \includegraphics[width=0.99\linewidth]{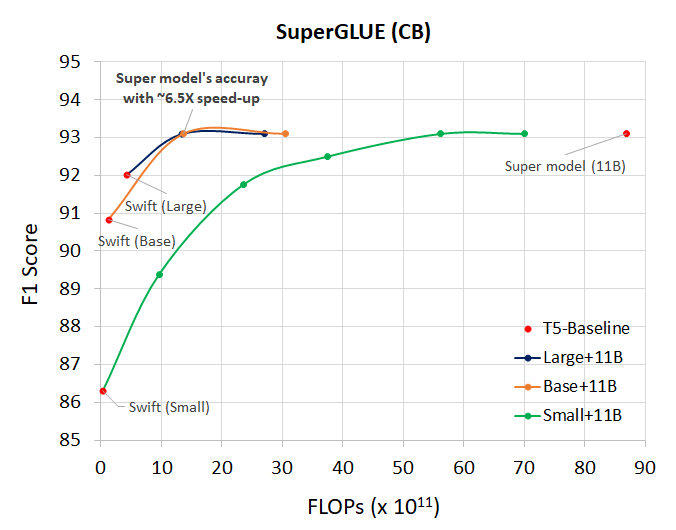}
    \end{subfigure}    
    \hspace{-10pt}
    \begin{subfigure}{0.49\textwidth}
        \centering
        \includegraphics[width=0.99\linewidth]{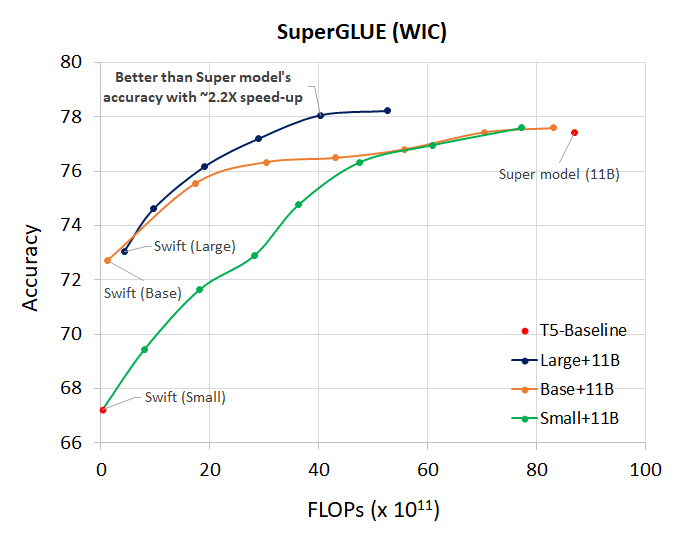}
    \end{subfigure}    
    \caption{\small Trade-off curves with T5 backbone on SuperGLUE tasks.}
    \label{fig:super_glue_curves_appendix}
\end{figure*}

The trade-off curves (for the experiments given in Table \ref{tbl:results}) with T5 architecture on GLUE and SuperGLUE tasks are respectively shown in Figures \ref{fig:glue_curves_appendix} and \ref{fig:super_glue_curves_appendix}. The ablation over different Swift models are also given in the figures.

\begin{figure*}[b!]
    \centering
    \begin{subfigure}{0.49\textwidth}
        \centering
        \includegraphics[width=0.99\linewidth]{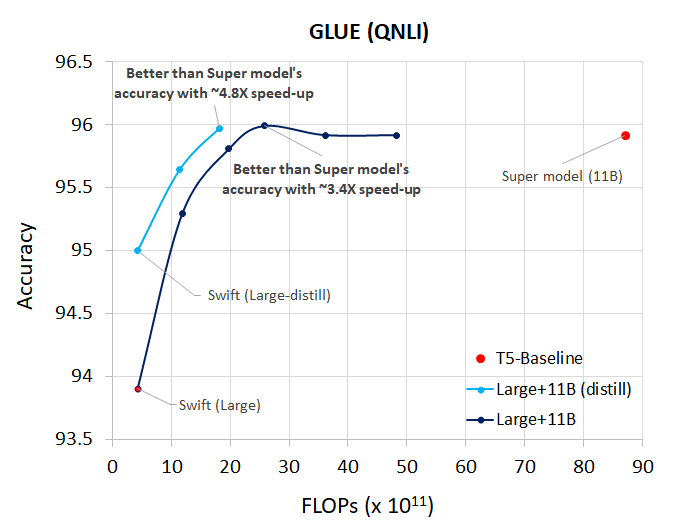}
    \end{subfigure}    
    \hspace{-10pt}
    \begin{subfigure}{0.49\textwidth}
        \centering
        \includegraphics[width=0.99\linewidth]{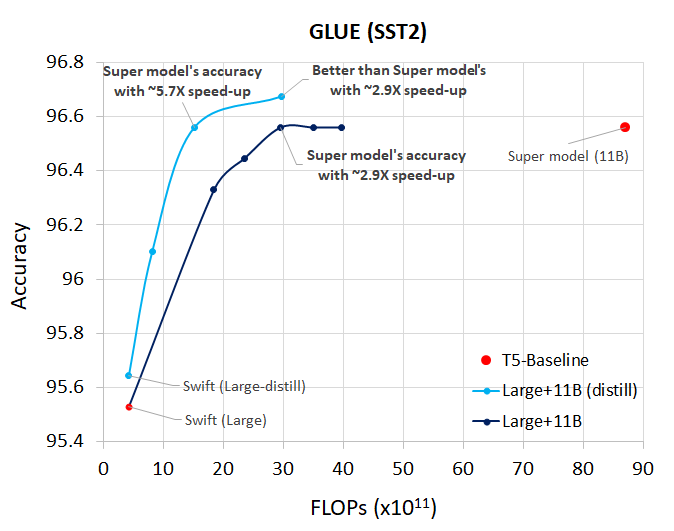}
    \end{subfigure}
    \hspace{-10pt}
    \begin{subfigure}{0.49\textwidth}
        \centering
        \includegraphics[width=0.99\linewidth]{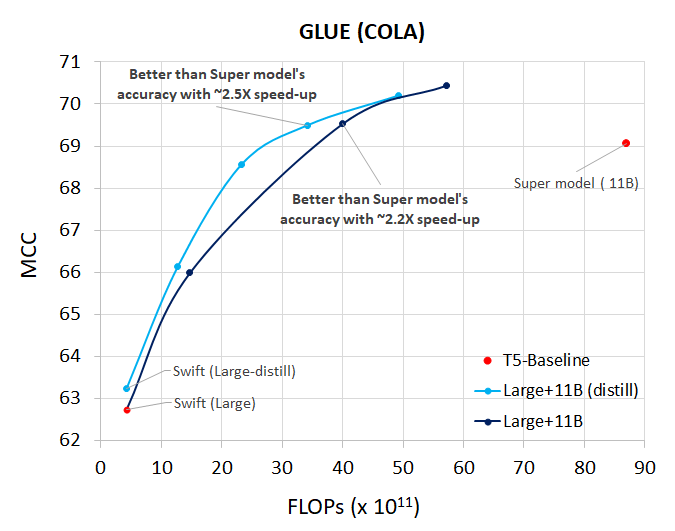}
    \end{subfigure}
    \caption{\small Distillation-based trade-off curves with T5 backbone on some GLUE tasks.}
    \label{fig:glue_curves_appendix_distill}
\end{figure*}

In Figure \ref{fig:glue_curves_appendix_distill}, the accuracy vs. FLOPs trade-off curves for distillation-based experiments (reported in Table \ref{tbl:t5_distill}) are also given. On QNLI, distillation-based E-LANG (denoted by DE-LANG) with 4.8$\times$ less computations than the Super model outperforms E-LANG with 3.4$\times$ FLOPs speed-up, although both methods performs 0.1\% more accurate than the Super model. DE-LANG on SST2 can also achieve the Super model's accuracy with 5.7$\times$ less computations, while the original E-LANG achieves the same performance with only 2.9$\times$ speed-up. Moreover, DE-LANG can improve the Super model's accuracy by 0.1\% with 2.9$\times$ speed-up on SST2. For COLA, DE-LANG achieves a better FLOPs speed-up of 2.5$\times$ than E-LANG with 2.2$\times$ speed-up, where both outperform the Super model's accuracy by 0.4\%. 

\begin{figure*}[b!]
    \centering
    \begin{subfigure}{0.49\textwidth}
        \centering
        \includegraphics[width=0.99\linewidth]{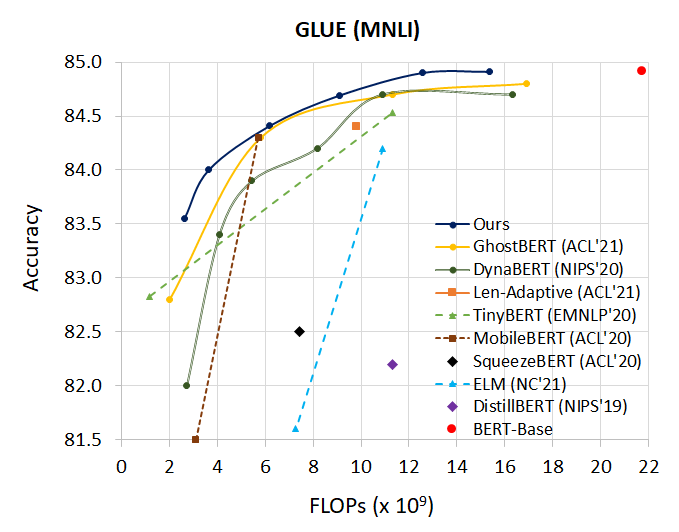}
    \end{subfigure}
    \begin{subfigure}{0.49\textwidth}
        \centering
        \includegraphics[width=0.99\linewidth]{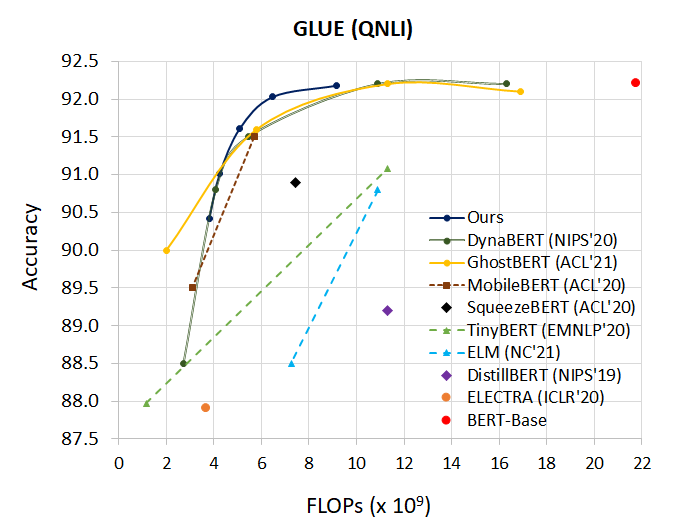}
    \end{subfigure}
    \begin{subfigure}{0.49\textwidth}
        \centering
        \includegraphics[width=0.99\linewidth]{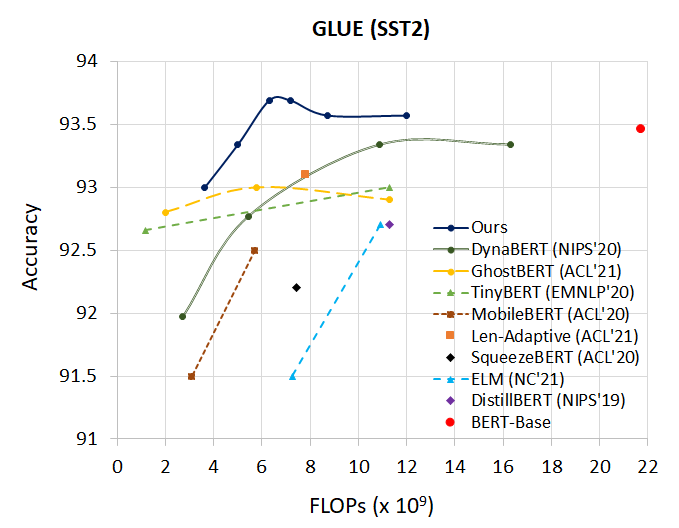}
    \end{subfigure}
    \begin{subfigure}{0.49\textwidth}
        \centering
        \includegraphics[width=0.99\linewidth]{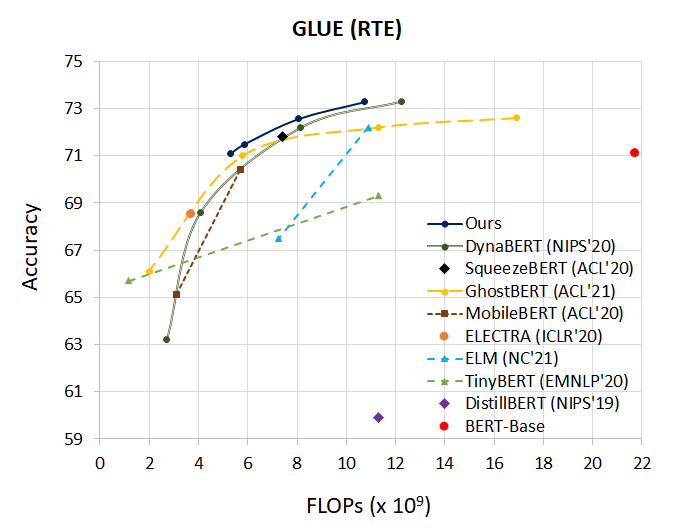}
    \end{subfigure}    
    \begin{subfigure}{0.49\textwidth}
        \centering
        \includegraphics[width=0.99\linewidth]{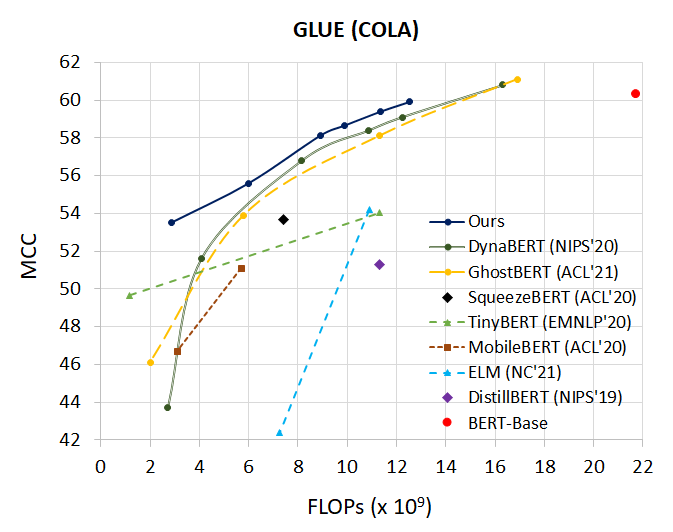}
    \end{subfigure}    
    \begin{subfigure}{0.49\textwidth}
        \centering
        \includegraphics[width=0.99\linewidth]{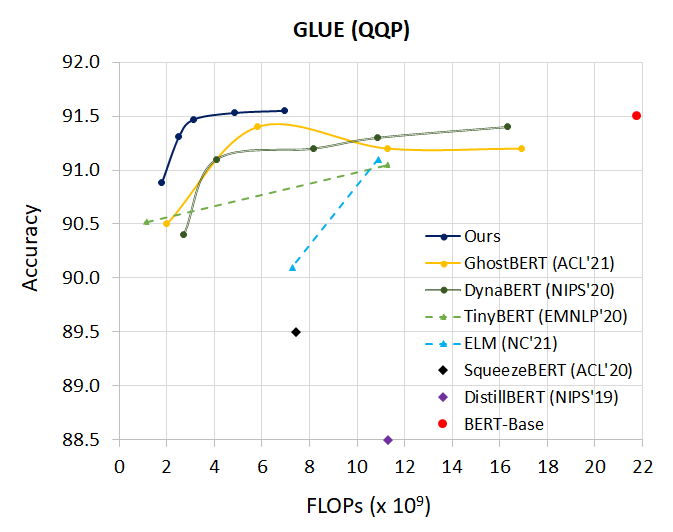}
    \end{subfigure}    
    \caption{\small Trade-off curves compared with BERT-based SOTA on GLUE tasks.}
    \label{fig:bert_glue_curves_appendix}
\end{figure*}

Figure \ref{fig:bert_glue_curves_appendix} also illustrates the corresponding curves for the BERT-based results of Table \ref{tbl:bert_results}, which are compared with previous works in fixed-size and adaptive inference.

\end{document}